# GCNBoost: Artwork Classification by Label Propagation through a Knowledge Graph

Cheikh Brahim El Vaigh
cheikh-brahim.el-vaigh@irisa.fr
Univ. Rennes, CNRS, IRISA
Lannion, France

Noa Garcia
noagarcia@ids.osaka-u.ac.jp
Osaka University, Institute for Databillity Science
Osaka, Japan

Benjamin Renoust
renoust@ids.osaka-u.ac.jp
Median Technologies, and Osaka University, Institute for Datability Science
Valbonne, France

Chenhui Chu
chu@ids.osaka-u.ac.jp
Kyoto University, and Osaka University, Institute for Datability Science
Osaka, Japan

Yuta Nakashima
n-yuta@ids.osaka-u.ac.jp
Osaka University, Institute for Datability Science
Osaka, Japan

Hajime Nagahara
nagahara@ids.osaka-u.ac.jp
Osaka University, Institute for Datability Science
Osaka, Japan

## Abstract

The rise of digitization of cultural documents offers large-scale contents, opening the road for development of AI systems in order to preserve, search, and deliver cultural heritage. To organize such cultural content also means to classify them, a task that is very familiar to modern computer science. Contextual information is often the key to structure such real world data, and we propose to use it in form of a knowledge graph. Such a knowledge graph, combined with content analysis, enhances the notion of proximity between artworks so it improves the performances in classification tasks. In this paper, we propose a novel use of a knowledge graph, that is constructed on annotated data and pseudo-labeled data. With label propagation, we boost artwork classification by training a model using a graph convolutional network, relying on the relationships between entities of the knowledge graph. Following a transductive learning framework, our experiments show that relying on a knowledge graph modeling the relations between labeled data and unlabeled data allows to achieve state-of-the-art results on multiple classification tasks on a dataset of paintings, and on a dataset of Buddha statues. Additionally, we show state-of-the-art results for the difficult case of dealing with unbalanced data, with the limitation of disregarding classes with extremely low degrees in the knowledge graph.

## 1 Introduction

Knowledge graphs (KGs), often used for content representation and retrieval, are powerful tools for multimedia data management e.g. [2, 9, 38]. They allow to model data as a set of entities (nodes) and the relations between them (edges). KGs can play a key role to support automatic systems developed to help preserving cultural heritage [1], such as classifying and retrieving historical newspapers [9], paintings [11], or Buddha statues [31]. Automatic artwork classification consists in classifying artworks according to attributes e.g. style, author, or time period [24, 26, 36].

A piece of art, beyond its visual aspect, bears a lot of contextual information (e.g. time, author), which plays an important role in defining the artwork itself (this is especially true for contemporary art that has narrower context than the classical art). Combining visual and contextual information has been shown to be a successful approach for artwork classification [11]. In [11], the contextual information is gathered in a KG to model the interactions between artworks and their attributes, which includes a semantic proximity that might not reside in the visual information itself. In our context, we combine the visual features of a given piece of art with its information embedded in an extended knowledge graph (EKG) that we define hereafter, with different digital archives.

Existing approaches for artwork classification are based on *inductive learning* [27] generalizing tons of observation, but limited by the cost of laborious human annotation. In contrast, transductive learning and label propagation [15, 44] can be used to learn from a smaller set and missing labels. Label propagation can predict pseudo-labels for unlabeled data (test data) and increase the amount of training samples at training time. KGs promote this transductive process [10, 15, 33] by modeling latent relations between labeled data and unlabeled data, facilitating for the same reason classification through labels propagation.

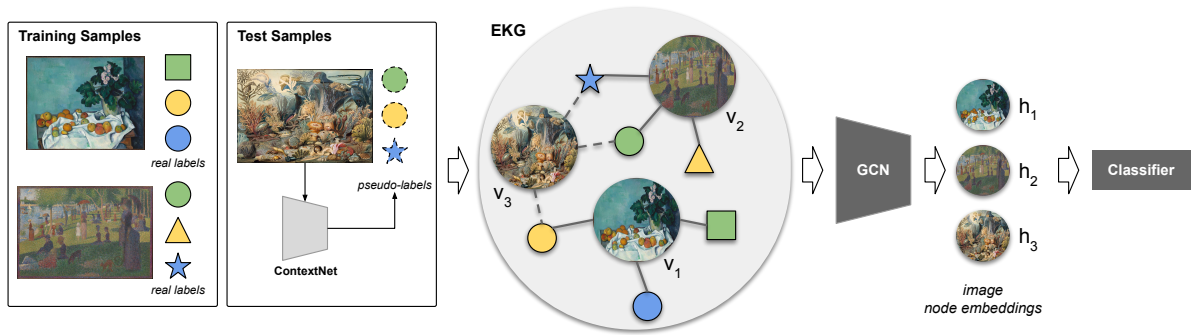

Figure 1: The overview of our proposed framework, named GCNBoost. The input are artworks with their labels (shapes) and unlabeled data that we pseudo-label with a pre-trained (state-of-the-art) model on artwork classification. Artworks and their labels are used to build an EKG that is fed into a GCN and the output embedding is used to build the final classifier. We do not show the initial embeddings of the EKG's nodes that are obtained with ResNet50 [14] for images and node2vec [13] for labels.

In this paper, we build our EKG based on a given set of entities (images) with their attributes (multiple labels) relying on labeled data as well as unlabeled data, to which we assign pseudo-labels. This EKG also captures relations between the different entities of the dataset. We learn embeddings for all nodes in the graph using a graph convolutional network (GCN). These embeddings are then used for artwork classification to predict multiple labels (i.e. multiple attributes) of a piece of art in a transductive learning framework (see Figure 1 for an overview). The proposed model is evaluated on the SemArt dataset [12], and the Buddha statues dataset [31], for eight different classification tasks, showing significant improvement with respect to previous work.

The main contributions of the paper are summarized as follows:
(1) We propose to build an EKG accounting for both labeled and unlabeled data by using preliminary assigned pseudo-labels.
(2) We devise a framework for digital art analysis that leverages GCNs to compute distinct artwork embeddings, which we show to be robust to unbalanced data.
(3) We evaluate our approach against state-of-the-art methods, obtaining higher accuracy on two different artistic domains: fine-art paintings and Buddha statues.

## 2 RELATED WORK

In this section we limit the discussion to the task of automatic art analysis (Section 2.1), and how it can be benefited from both image classification with GCNs (Section 2.2) and label propagation (Section 2.3).

### 2.1 Automatic Artwork Classification

Historically, the task of automatic art analysis was initially addressed using handcrafted features to describe the visual content of a digitized artwork [5, 16, 26]. Those handcrafted features ranged from color [42] or brushwork [19] detection to scale-invariant local features [32], and were used to classify pieces of art according to their attributes[1] through SVM classifiers. However, those approaches were bounded by the quality of the features themselves.

With the emergence of machine learning techniques that automatically extract features from an image using pre-trained convolutional neural networks (CNNs), such as ResNet [14] or VGG [34], the need for handcrafted features was replaced. Pre-trained CNNs could capture accurate representations for different kinds of entities, such as text, natural images, or art pieces, and thus, they were extensively used as an off-the-shelf method to classify artworks [3, 11, 12, 23, 24, 31]. CNNs have also been fine-tuned to devise multitask art classifiers [11, 23, 31, 36]. Meanwhile, features extracted from CNNs only contain information about the visual aspect of the image, without considering the cultural and historical context of the artworks. To skirt this issue, the authors of [12] proposed to use a joint visual and textual representation for fine-art paintings, allowing a multimodal analysis and opening the door to study art from the semantic point of view.

To further study art from the semantic perspective, visual information can be complemented with specific knowledge about the art pieces, such as the artists, the period of time, or the style. This allows to incorporate the general context of the artworks, such as the social and historical context, into their representation. For instance, in order to have contextual representations of artworks and their attributes, a multi-task learning approach is developed in [11], allowing different paintings to interact through their attributes. Moreover, KGs can be used to build an accurate representation [11, 31] of artworks and their attributes based on KG embedding models such as node2vec [13]. The latter is used in [11, 31] jointly with deep visual features, showing state-of-the-art results of painting and Buddha statues classification. Our paper follows the same direction by devising a semantic KG for art analysis. Furthermore, we study how unlabeled data can be used within a transductive setup, in order to improve the quality of artworks' interrelationships in a KGs, facilitating automatic art analysis.

### 2.2 Image Classification with GCNs

In the recent work on image classification, GCNs are receiving more attention thanks to their ability to model the relationships between a set of entities through a KG, which is effective in the tasks of node classification [28, 35] and link prediction [35]. Particularly, GCN are used to model the relations a set of images may have accounting for their labels [6, 22, 39, 43]. The basic idea is to combine the classical visual features with GCN embeddings learned over the KG. For example, GCNs are used in [6] to improve multi-labels classification accounting for semantic links between different labels, whereas the authors of [39] improved this idea with WordNet concepts hierarchy, devising a zero-shot classification technique. Finally, the authors in [43] used a weighted adjacency matrix to efficiently model the inter-dependency between image labels. We follow the same idea, introducing pseudo-labeled data, used as true data to train a GCN following a standard label propagation process. However, training GCNs with noisy data is challenging, as they are based on the hypothesis of an isotropic averaging operation, meaning that pseudo-labeled data have the same influence to their neighborhood as the ground truth. To alleviate this issue, we use different pseudo-labels for each artwork.

---
[1]e.g. author, style movement, or period of time.

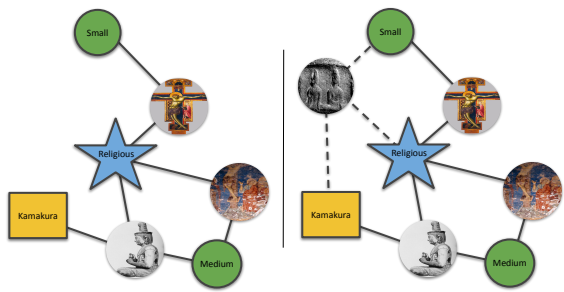

**Figure 2: An example of a KG and its EKG. Each node corresponds to either a painting, a Buddha statue or an attribute. The plain lines correspond to existing relationships in the KG (left), while dashed lines are obtained with pseudo-labels to build the EKG (right).**

## 2.3 Label propagation

Label propagation over networks has been a successful strategy to help classification of pseudo-labeled data [29]. It has recently been used for transductive learning with GCNs to incrementally label test data [8, 15]. The authors in [8] used online pseudo-labeling process for unlabeled data, while the authors in [15] performed the pseudo-labeling offline without changing it at learning time. Our work is a combination of the two. We start with pseudo-labels produced by a pre-trained model, and we refine those pseudo-labels during training.

## 3 Approach

Our task is defined as a multi-label classification problem. Formally, given an artwork $x$, we predict a set of its associated labels $\{t_c | c \in C\}$, where $t_c$ is a label in a certain set $L_c$ of labels and $C$ is the set of label category indices. Taking the SemArt dataset [12] as an example, we have *Type*, *School*, *TimeFrame*, and *Author* in $C$, and each set has specific labels, i.e., *portrait* in $L_{type}$, *Dutch* in $L_{school}$, and *Vincent van Gogh* in $L_{author}$. A straightforward approach to solve this problem is to adopt a multi-task learning framework with a CNNs, in which a single feature extractor is shared among classifiers dedicated for $L_c$'s.

As aforementioned, attributes (or corresponding labels) are rarely independent to each other in a real-world multi-label dataset. A relationship between two labels may be provided as auxiliary knowledge in the dataset or can be extracted from external knowledge sources, such as DBpedia[2]. Such relationships include the time frame of an artwork and its author, stating that the author lived in that period of time (overlapped over the same period of time). For example, the SemArt dataset [12] comes with such auxiliary information that links between labels: *Vincent van Gogh* is from the *Dutch* painting school, and all his painting have the same school, which can be represented by an edge connecting author *Vincent van Gogh* to school *Dutch*. Such knowledge tells possible correlations among artworks, which can facilitate image representation.

This leads us to reformulating our multi-label classification task with the transductive learning paradigm, explicitly modeling the correlations among labels and artworks in both training and test sets through a KG as in Figure 2. Test images have no labels or missing labels; we thus assign *pseudo-labels* to make noisy connections in the graph and compute an embedding of each image. This allows to capture different relations between art pieces. That is, when two artworks are from the same author, they are semantically related and connected through a path of length two, and should be closer when compared to non-related artworks (with regards to author nodes).

This transductive learning paradigm, as shown in Figure 1, is core to our classification framework (named GCNBoost): we create an extended knowledge graph (EKG) based on pseudo labels predicted by a model on a training set, then infer multiple artwork classes using a graph convolution network (GCN). The following subsections describe our KG construction, image embedding using the EKG, and training and inference process with pseudo-labels.

### 3.1 KG Construction

A KG is a graph $\mathcal{G} = (\mathcal{V}, \mathcal{E})$, in which the entities and their relations are represented by a collection of nodes $\mathcal{V}$ and edges $\mathcal{E}$, respectively. Nodes in $\mathcal{V}$ are the artworks and their labels (i.e., attributes). The latter can be authors—the entities that created the artworks—or types—categories of art such as portraits or landscapes—of the artworks. The edges in $\mathcal{E}$ represent the relations between two entities in $\mathcal{V}$. The semantic of those relations depend upon the underlying entities. For example, an edge between a certain artwork and a certain author represents the artwork is created by the author. Thus, the KG captures the contextual knowledge and the semantics of the relationships between the artworks and their labels.

Formally, let $X_{\text{train}}$ and $X_{\text{test}}$ denote the sets of artworks for training and test, respectively. Artwork $x \in X_{\text{train}}$ is associated with multiple labels $\{t_c | c \in C\}$. The assignment of a label to an artwork gives edge $(x, t_c)$, and the set of all edges are denoted by $\mathcal{W}$. As mentioned above, some auxiliary sources of knowledge can provide explicit relationships among labels $\mathcal{K} = \{(l_i, l_j) | l_i, l_j \in \mathcal{L}\}$, where $\mathcal{L} = \bigcup_c L_c$. Using these definitions, sets $\mathcal{V}$ and $\mathcal{E}$ can be

---
[2]https://wiki.dbpedia.org/

given by
$$\mathcal{V} = \mathcal{X}_{\text{train}} \cup \mathcal{L} \quad (1)$$
$$\mathcal{E} = \mathcal{W} \cup \mathcal{K}. \quad (2)$$

For transductive learning, we extend the KG $\mathcal{G}$ with the test set $\mathcal{X}_{\text{test}}$. Let $\mathcal{G}' = (\mathcal{V}', \mathcal{E}')$ denote our extended knowledge graph (EKG). $\mathcal{V}'$ is a simple extension of $\mathcal{V}$ with the test set, that is,
$$\mathcal{V}' = \mathcal{V} \cup \mathcal{X}_{\text{test}}. \quad (3)$$
Artworks in the test set have no associated labels and so no edges. In order to facilitate the learning of image embeddings, we use an initial guess of labels for the test set, namely pseudo-labels, and add them to the graph. Let $t'_c = g_c(x)$ denote the label in $L_c$ predicted by a (separately) trained classifier $g_c$ for $x$ in test set $\mathcal{X}_{\text{test}}$. We can add this to $\mathcal{E}$ as new nodes and $\{(x, t'_c)\}$ as new edges for all $x \in \mathcal{X}_{\text{test}}$ and all $c \in C$. We thus have our EKG $\mathcal{G}' = (\mathcal{V}', \mathcal{E}')$, where
$$\mathcal{E}' = \mathcal{E} \cup \{(x, t'_c) | x \in \mathcal{X}_{\text{test}}, c \in C\}. \quad (4)$$

### 3.2 Image Embedding with EKG

We adopt a GCN [18] to encode into artwork (and label) embeddings the relationships provided by $\mathcal{G}'$. Our GCN has multiple layers, and the $n$-th layer's ($n = 1, \ldots, N$) hidden state $H^{(n)}$ is given using $(n-1)$-th layer's $H^{(n-1)}$ by
$$H^{(n)} = \text{ReLU}(D^{-\frac{1}{2}}(A+I)D^{-\frac{1}{2}} H^{(n-1)} W^{(n)} + b^{(n)}). \quad (5)$$
$A \in \mathbb{R}^{|\mathcal{V}'| \times |\mathcal{V}'|}$ is the adjacency matrix derived from $\mathcal{G}'$, $D$ is a diagonal matrix of node degrees given by $D_{ii} = \sum_j A_{ij}$, and $I$ is the identity. $W^{(n)} \in \mathbb{R}^{d \times d}$ is a matrix of learnable parameters. $H^{(0)} \in \mathbb{R}^{|\mathcal{V}'| \times d}$ is a stack of initial features of nodes, where row vector $h_i^{(0)}$ corresponds to the feature vector for node $v_i$. For images, we use ResNet50 [14] to obtain the initial feature vector. For labels, initial feature vectors are given by node2vec [13] over $\mathcal{G}'$.

With this GCN architecture, we can learn latent relationships between the nodes of the EKG. The adjacency matrix A provides direct relationships among the nodes in $\mathcal{G}'$, and the GCN propagates the information according to the edges. This process is particularly efficient for label nodes that serve as hub—nodes with high degree—as they will be equally important to their neighborhoods and facilitate the task of classification, by short cutting the path between images and labels nodes. For instance, we directly use the relationships between artwork and label nodes, or between label nodes themselves as we can easily interpret such relationships. Meanwhile, the model relies on nodes relationships in general such as indirect relationships through hubs.

### 3.3 Training and Inference

Given the artwork embedding $h_i = h_i^{(N)}$ for the node $v_i \in \mathcal{X}_{\text{train}} \cup \mathcal{X}_{\text{test}}$, a fully connected (FC) classifier $f_c$ with softmax makes prediction for each set $L_c$ of labels. Formally,
$$y_{ic} = f_c(x_i) = \text{softmax}(W_c h_i + b_c), \quad (6)$$
where $W_c \in \mathbb{R}^{|L_c| \times d}$ and $b_c \in \mathbb{R}^{|L_c|}$ are learnable parameters. The loss function $\ell$ is given by
$$\ell = - \sum_{x_i \in \mathcal{X}_{\text{train}}} \sum_{c \in C} \sum_{k=1}^{|L_c|} t_{ick} \log y_{ick}, \quad (7)$$

where $t_{ick}$ is the $k$-th element of the one-hot vector of label $t_{ic}$ for $x_i$. The loss function is evaluated only over the training set $\mathcal{X}_{\text{train}}$. After training, the $h_i$ corresponding to image $x_i \in \mathcal{X}_{\text{test}}$ is fed to the $f_c$ and softmax to predict its label.

## 4 Experiments

We conducted a thorough experimental evaluation of our framework on both the SemArt and Buddha statues datasets. We showcase performance against state of the art, and robustness with respect to noise in the relationships between nodes of a given graph. After the description of the datasets used in our experiments in Section 4.1, we present our implementation details in Section 4.2. The ablation studies on our approach are gathered in Section 4.3, comparing the importance of using different attributes of art pieces when building the KG as described in Section 3.1. Finally, we compare our classification approach to state-of-the-art competitors, showing significant improvement on the SemArt painting dataset in Section 4.4, and the Buddha statues dataset in Section 4.5.

### 4.1 Evaluation Datasets

To evaluate our artwork classification model, we used two datasets: SemArt [12] and Buddha statues [31].

**SemArt dataset** The SemArt dataset contains 19,244 train, 1,069 test, and 1,069 validation images of European fine-art paintings. Each painting is labeled with the attributes *Author*, Title, Date, Technique, *Type*, *School*, and *TimeFrame*. We use the pre-trained ContexNet model [11] to produce the initial pseudo-labels for the test samples. Following previous work [11], we evaluate our models on four classification tasks:

- **Type classification (T1.1)** consists in classifying each painting in one of the 10 following Types: portrait, landscape, religious, study, genre, still-life, mythological, interior, historical and other.
- **School classification (T1.2)** aims at assigning a class to each painting from one of the 24 following classes: Italian, Dutch, French, Flemish, German, Spanish, English, Netherlandish, Austrian, Hungarian, American, Danish, Swiss, Russian, Scottish, Greek, Catalan, Bohemian, Swedish, Irish, Norwegian, Polish, Other and Unknown.
- **TimeFrame classification (T1.3)** targets classification of the painting according to 17 different period of time and one additional class (*Unknown*) for or TimeFrames that are not considered classes. Each class contains at least 10 training samples.
- **Author classification (T1.4)** performs painting classification according to 350 different painters. We first do so by including all the authors (although some authors have only one training sample in the dataset). In a second experiment, we filtered authors so that we keep only the ones with at least ten paintings in the training set.

**Buddha statues dataset** Similarly, the Buddha statues dataset contains 2,665 samples split into 1,866 train, 533 test, and 266 validation, with the attributes *Style*, *Size*, *Century of creation*, and *Dimensions*. To compute the pseudo-labels, we trained the proposed neural network (NN) model from [31], which we also use as a baseline. We evaluate our model on the four tasks defined in [31]:

- **Statue style classification (T2.1)** consists in classifying Buddha faces in one of this four different styles: China, Kamakura period, Nara period, and Heian period.
- **Statue size classification (T2.2)** assigns a class to each Buddha faces from one of this three different styles: Small, Medium and Big.
- **Statue century of creation classification (T2.3)** aims at classifying Buddha faces into seven different centuries: V, VI, VII, VIII, IX, XII, XIII. We disregard centuries with few training samples.
- **Statue dimensions classification (T2.4)** performs Buddha faces classification according to 12 different dimensions ranging from 50 to 1,050 cm.

## 4.2 Implementation Details

At learning time, we use the whole dataset at each iteration, allowing a global optimization without mini-batches. While global training is memory expensive in general, since we have relatively small datasets that fit in memory, it was more convenient and faster than using mini-batches, specially when training on GPUs.

For artwork classification, we train a two-layer GCN as described in Section 3.2 for node classification. Node embeddings are initialized using node2vec for training and validation samples, and randomly for test samples. The dimension of this embedding is 128. The first layer of our GCN takes the initial embedding, and outputs a hidden representation of size 16, following the GCN hyper-parameters [18]. This first level GCN embedding is introduced in the second layer which an output size equals to the number of labels $|L_C|$. We train a separate model for each task.

We use Adam [17] for gradient optimization with the learning rate set to 0.001, and a maximum number of iterations set to 2,000. We also implement an early stop mechanism with a window size of 100, i.e. we stop training if the validation loss does not decrease for 100 consecutive iterations.

## 4.3 Importance of Pseudo-Labels Assignment

We first evaluate on the SemArt dataset the effect of assigning pseudo-labels to the test set to build our proposed EKG.

**Baselines** For comparison, we use two models: the original ContextNet [11], and our proposed model based on EKG but with a random assignment of pseudo-labels ($S_0$ Random initialization).

**Pseudo-label assignment** In the rest of our models, we initally predict pseudo-labels using the pre-trained ContextNet. To analyse its contribution, we add between one to four categories of pseudo-labels at a time, referred as models $S_1$, $S_2$, $S_3$, and $S_{all}$, respectively.

Results are shown in Table 1. Rows 1 and 2 report the accuracies of the baselines, showing that 1) ContextNet [11] is a strong model, and 2) the random initialization does not work at all.

Rows 3 to 6 show results when adding only one category of pseudo-labels at a time ($S_1$). This strategy improves the classification accuracy over the original ContextNet on *School* by 4 points and *TimeFrame* by 18 points. Results are close to the baseline on *Type*. However, we lose 42 points on *Author*, mostly because of the unbalanced data in this category (further details in Section 4.4). Moreover, $S_1$ strategy only improves accuracy for the categories used to build the graph, obtaining poor results on the others. For example, using *School* pseudo-labels (Row 4) improves only the results for the same category, *School*, and obtains poor results on *Type*, *TimeFrame*, and *Author*.

Table 1: Classification results on SemArt. The first part of this table gives the baseline. The following groups give the classification accuracy using one, two, three, then all categories of pseudo-labels to build the KG.

|    | Model | Type T1.1 | School T1.2 | TimeFrame T1.3 | Author T1.4 |
|----|-------|-----------|-------------|----------------|-------------|
| 1  | ContextNet [11] | 0.815 | 0.671 | 0.613 | **0.615** |
| 2  | $S_0$ Random intialization | 0.100 | 0.010 | 0.050 | 0.040 |
| 3  | $S_1$ Type | 0.807 | 0.368 | 0.136 | 0.013 |
| 4  | $S_1$ School | 0.296 | 0.718 | 0.154 | 0.035 |
| 5  | $S_1$ TimeFrame | 0.363 | 0.392 | 0.796 | 0.047 |
| 6  | $S_1$ Author | 0.352 | 0.420 | 0.271 | 0.181 |
| 7  | $S_2$ School_Author | 0.397 | 0.866 | 0.314 | 0.284 |
| 8  | $S_2$ School_Type | 0.861 | 0.739 | 0.213 | 0.058 |
| 9  | $S_2$ School_TimeFrame | 0.363 | 0.846 | 0.843 | 0.139 |
| 10 | $S_2$ Type_Author | 0.899 | 0.486 | 0.295 | 0.302 |
| 11 | $S_2$ Type_TimeFrame | 0.915 | 0.485 | 0.853 | 0.098 |
| 12 | $S_2$ Author_TimeFrame | 0.417 | 0.504 | 0.906 | 0.354 |
| 13 | $S_3$ School_Author_TimeFrame | 0.461 | 0.882 | **0.933** | 0.435 |
| 14 | $S_3$ Author_TimeFrame_Type | 0.930 | 0.564 | 0.932 | 0.482 |
| 15 | $S_3$ Author_School_Type | 0.924 | 0.827 | 0.348 | 0.394 |
| 16 | $S_3$ School_Type_TimeFrame | 0.929 | 0.859 | 0.877 | 0.204 |
| 17 | $S_{all}$ | **0.939** | **0.889** | 0.927 | 0.479 |

Rows 7 to 12 show results when two categories of pseudo-labels are used to build the EKG ($S_2$). Results are improved with respect to $S_1$ on multiple ways. First, accuracy is enhanced for the considered categories: e.g. using *School* and *Type* (Row 8) raises the accuracy for both categories over $S_1$ when only using *School* (Row 4) or *Type* (Row 3). Second, accuracy is also improved for the other categories, e.g. *TimeFrame* and *Author* are better in Row 8 than in Row 3 or Row 4. This gain in accuracy highlights the interest of using existing nodes interrelationships to capture hidden relations between nodes, which leads to improve the accuracy for the classification task. Third, from a qualitative point of view, we observe that the results for some pairs of pseudo-labels are coherent with the way we create the EKG. For example, *Type* and *TimeFrame* (Row 11) obtain the best scores among all the $S_2$ models because a large number of painting types were famous in a particular period of time, which creates a hub of nodes around *TimeFrame*, and *Type*.

Rows 13 to 16 show results combining three categories of pseudo-labels ($S_3$), whereas Row 17 provides accuracies when all the categories are used ($S_{all}$). These strategies further improve results, observing similar conclusions as with strategy $S_2$. Meanwhile, $S_3$ is better than $S_{all}$ on *TimeFrame*. While the difference in accuracy is only one point, this suggests that adding extra information to the EKG may incorporate too much noise. Thus, adding three categories of pseudo-labels seems to always improve the accuracy for the SemArt dataset, while four categories may or may not improve the results.

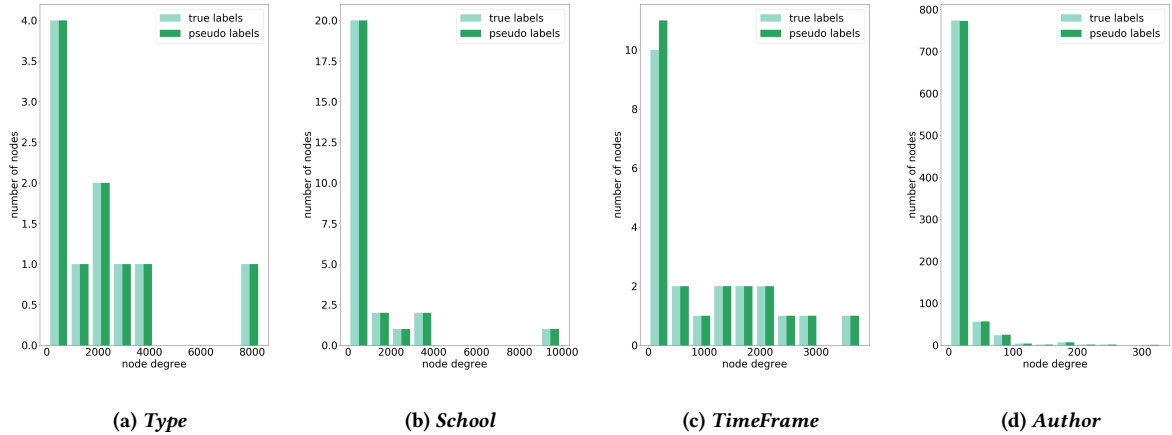

Figure 3: Distribution of nodes degrees for the four categories (*Type*, *School*, *TimeFrame*, *Author*) based on the true classes for test and train data, and the pseudo-labels for the test.

Table 2: Comparison of different classification results on SemArt. The row GCNBoost $S_1$ to $S_{all}$ gives different configurations of our classification approach

| Model | Type T1.1 | School T1.2 | TimeFrame T1.3 | Author T1.4 |
|---|---|---|---|---|
| VGG16 pre-trained [11] | 0.706 | 0.502 | 0.418 | 0.482 |
| ResNet50 pre-trained [11] | 0.726 | 0.557 | 0.456 | 0.500 |
| ResNet152 pre-trained [11] | 0.740 | 0.540 | 0.454 | 0.489 |
| VGG16 fine-tuned [11] | 0.768 | 0.616 | 0.559 | 0.520 |
| ResNet50 fine-tuned [11] | 0.765 | 0.655 | 0.604 | 0.515 |
| ResNet152 fine-tuned [11] | 0.790 | 0.653 | 0.598 | 0.573 |
| ResNet50+Attributes [11] | 0.785 | 0.667 | 0.599 | 0.561 |
| ResNet50+Captions [11] | 0.799 | 0.649 | 0.598 | 0.607 |
| ContextNet MTL [11] | 0.791 | 0.691 | 0.632 | 0.603 |
| ContextNet KGM [11] | 0.815 | 0.671 | 0.613 | 0.615 |
| GCNBoost $S_1$ | 0.807 | 0.718 | 0.796 | 0.181 |
| GCNBoost $S_2$ | 0.915 | 0.866 | 0.906 | 0.354 |
| GCNBoost $S_3$ | 0.930 | 0.882 | **0.933** | 0.482 |
| GCNBoost $S_{all}$ | **0.939** | **0.889** | 0.927 | 0.479 |
| GCNBoost $S_{all}^*$ (Author filter) | - | - | - | **0.702** |

### 4.4 SemArt Evaluation

Next, we compare our approach to a series of methods that reported state-of-the-art results on the SemArt dataset, namely:

- **Pre-trained Networks [11].** VGG16 [34], ResNet50 [14], and ResNet152 [14] trained for natural image classification.
- **Fine-tuned Networks [11].** VGG16, ResNet50, and ResNet152 fine-tuned on the SemArt dataset.
- **ResNet50+Attributes [11].** Combination of ResNet50 features with the predicted attributes by one of the three previous fine-tuned models.
- **ResNet50+Captions [11].** Combination of ResNet50 features with a generated caption embeddings [41].

- **ContextNet models [11].** ContextNet includes a multi-task learning model (ContextNet MTL), which learns the classification of the four categories of a painting as a four shared tasks, and a knowledge graph model (ContextNet KGM), which is the one used in this work to predict the pseudo-labels. ContextNet KGM builds and uses an artistic KG to model relation between nodes and to perform classification, combining features extracted with a ResNet50 [14] for visual features and the model node2vec [13] for nodes embedding.

The models above can be divided into two groups: 1) local, disregarding the context (Pre-trained Networks, Fine-tuned Networks, and ResNet50+Captions); and 2) global, incorporating as much context as possible (ResNet50+Attributes, ContextNet). The scores for the different methods are obtained from [11], except for ContextNet that are computed using the pre-trained models on Github.[3]

Results are reported in Table 2. The proposed GCN classification strategy ($S_{all}$) outperforms previous methods by a large margin in all the categories except *Author*, where the accuracy is very low compared to ContextNet. One can note that ContextNet MTL and ContextNet KGM are complementary on the four categories of attributes, as they model the global nodes interrelationships differently, but they also obtain low scores on *TimeFrame* and *Author*. This can be explained by the imbalance in the number of training samples in those categories (especially for *Author*). While our GCNBoost model is able to recover the problem for *Type*, *School* and *TimeFrame* (they have a only small unbalancing), the same approach failed with *Author*.

To understand the imbalance problem for the *Author* category, in Figure 3, we plot the degrees distribution of the EKG nodes for the different attributes (one sub-plot per attribute). We can see that all the attributes are imbalanced. However, the attributes in *Author* (Figure 4d) present a huge imbalance: there are around 2, 000 *Author* nodes with small connectivity (small number of associated paintings) and less than 200 nodes with medium to high connectivity (more than 50 associated paintings). To show how this data imbalance affects the result in our proposed model, we disregard *Author* nodes with a degree less than five (i.e. to have at least 10 training examples). The result for this setup is given as GCN $S_{all}^*$ *(Author filter)* (bottom line of Table 2). We can see that by removing painting data points with few learning samples, we are able to improve the accuracy. However, the last result is not directly comparable with the rest of the approaches in Table 2 since we trimmed the training set to have enough training samples.

---
[3]https://github.com/noagarcia/context-art-classification

Table 3: Classification results on Buddha dataset. $S_{all}$ is using style, size, century, and dimensions to create the EKG

|   | Model | Style T2.1 | Size T2.2 | Century T2.3 | Dimensions T2.4 |
|---|---|---|---|---|---|
| 1 | NN (*original*) [31] | **0.98** | 0.78 | 0.78 | 0.78 |
| 2 | NN (*retrained*) | 0.58 | 0.65 | 0.76 | 0.46 |
| 3 | $S_0$ Random intialization | 0.23 | 0.30 | 0.13 | 0.08 |
| 4 | GCNBoost $S_1$ | 0.57 | 0.68 | 0.74 | 0.47 |
| 5 | GCNBoost $S_2$ | 0.59 | 0.85 | 0.80 | 0.76 |
| 6 | GCNBoost $S_3$ | 0.88 | 0.86 | 0.86 | 0.84 |
| 7 | GCNBoost $S_{all}$ | 0.92 | **0.94** | 0.93 | **0.90** |

### 4.5 Buddha Statues Evaluation

We apply our approach on the Buddha statues dataset following the same process as for the SemArt dataset. We use as a baseline the NN classifier defined in the original work [31]. It is a fully connected layer followed by a softmax activation with categorical cross entropy loss based on image features (the best results in the paper are given with ResNet-50 features [31]). This model is used in two setups:

- **NN original** is exactly the same architecture and setup as given by the authors of the original paper [31]. It is trained on a set of 3,334 images; however some images do not have all of the four different attributes (sparse data). This setup uses a 70%-30% split of the dataset with 5-fold cross validation, and does not keep sample for a test set.
- **NN retrained** is also the same architecture as given by the authors of the original paper [31], but with a different setup, that is common to all other experiments for Buddha statues. We consider a 70%-20% split of the Buddha statues dataset for train and validation, and keep the 10% of the remaining statues as an independent test set. Furthermore, each statue we consider bears all of the four attributes of our tasks, with a total of 2,665 images.

We build the EKG in the same spirit as for SemArt, but of course using different attributes. We leverage on the NN model (retrained) to build the pseudo-labels. We show in Figure 4 the node distributions for this EKG, and notice some imbalance (although less than for the SemArt dataset). Our GCNBoost shows robustness to this issue, by mitigating the different attributes of a statue as we discussed in the previous section.

To further investigate the impact of each attribute, we use the same strategy of adding one at a time. The results are gathered in Table 3. The first row gives the best results reported for the NN of the original paper [31]. The second row gives our baseline with the retrained NN on our experimental setup, while the following rows give our GCNBoost model for each different strategy. The results of the NN models greatly vary across the different attributes, even in the best reported case. In contrast, GCNBoost is able to improve most of the classification results, in spite of the limited available data (only 2,665 samples). Moreover, we also notice a consistent accuracy increase from the worst model, $S_0$ with random pseudo-labels assignment, to the best model, $S_{all}$ that includes all artwork relationships.

Finally, the original reported NN results still show a better accuracy on *Style* classification *(T2.1)*. This may be explained by the fact that it uses a 5-fold cross validation, with more training data as compared to our protocol. GCNBoost and the retrained NN baseline retain 10% of artworks to form a separated test dataset (that was not done in the original paper). This difference of setup might explain the difference between performances of the two experiments.

## 5 Discussion

In order to understand the quality of our approach with the EKG and GCN, we further investigate the role of node degree distributions in the EKG, as well as the ability of GCNBoost to cluster the different art pieces while sharing similar attributes just as homophily in real world networks.

Homophily [4] is a widely studied concept in complex networks. It follows the intuition that the more entities share attributes, the more they tend to assemble together (for example, people in social ties [25]), bringing cohesion to clusters of entities [30]. Combining visual features and the KG builds upon this idea: visually similar paintings have a high chance to share some semantic properties. While the generation of pseudo-labels introduce some level of noise, the GCN mitigates this noise by considering neighborhood at a larger level, comparable to a smoothing process. This effect however comes to a limitation when information is insufficient. This is due to the long-tail distribution of node degrees, typical of real-world data [7]: a few observations tend to be extremely well connected, while many observations are little to not connected at all. As a consequence, we can observe a very imbalanced distribution of labels, and the *Author* class makes one good example (see Figure 4d). It thus becomes difficult, from the graph perspective, for a random walk to reach authors with a low degree (i.e. a low number of paintings), and then predict them to any given painting. Investigating different loss functions (such as focal loss [20] or distribution-balanced loss [40]) may help mitigate this issue.

We further visualize using tSNE [37] the embeddings of ContexNet (Figure 5a) and GCNBoost (Figure 5b) learned when predicting the category *TimeFrame* on SemArt, as well as the embeddings of the baseline NN model (Figure 5c) and GCNBoost (Figure 5d) when predicting the category *Style* on Buddha dataset. Following the idea of homophily [4] we can see that ContexNet (Figure 5) globally clusters well the different timeframe categories but still displays some noise in between the clusters. The Resnet50 based

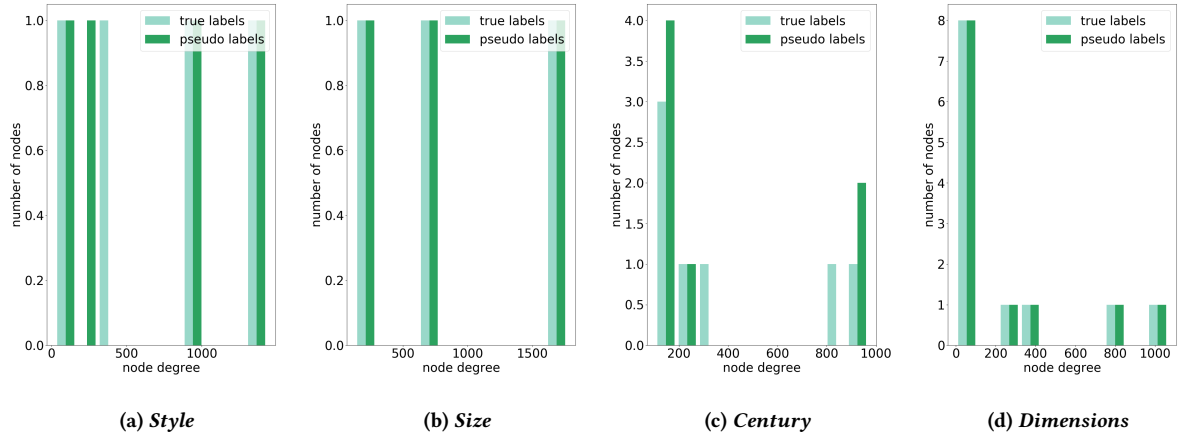

**Figure 4: Distribution of nodes degrees for the four categories (*Style, Size, Century, Dimensions*) based on the true classes for test and train data, and the pseudo-labels for the test.**

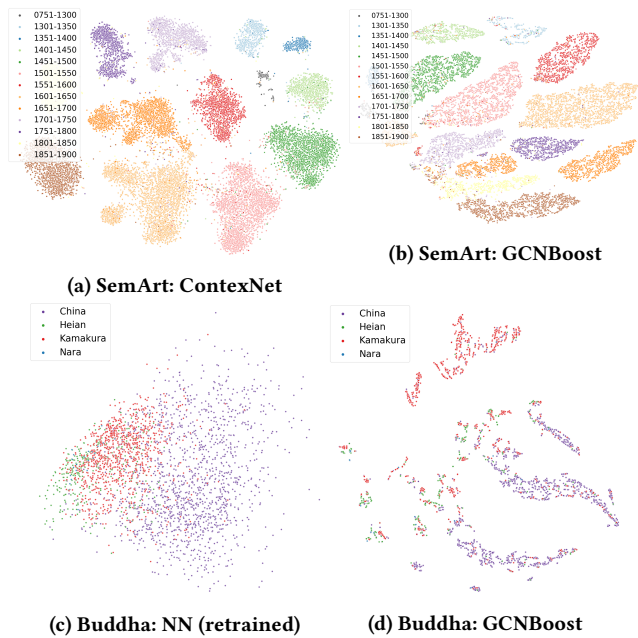

**Figure 5: Visualization using tSNE [37] of the embeddings of ContextNet [11] (a) and our GCNBoost model (b), learned when predicting the category *TimeFrame* on SemArt [12], and the NN [31] (c) and our GCNBoost (d) on Buddha statues [31].**

NN model puts everything in the same cluster (Figure 5c). In contrast, GCNBoost (Figure 5b and Figure 5d) tends to display better separated clusters (but not only limited to separating artworks based on their timeframe or their style). We keep for future work the investigations that might reveal the combinations of visual and semantic attributes explaining the clusters.

## 6 Conclusion and Perspectives

This paper shows a method that increases the effectiveness of the use of a knowledge graph (KG) in the context of multi-label classification with a transductive learning framework. Our method leverages on the inclusion of unlabelled data, which get pseudo labels attributed through label propagation enriching the KG, forming an extended KG (EKG). Using graph convolution networks (GCN) on the EKG, we have shown experimentally improvement on different classification tasks of two different artwork datasets (SemArt [12] and Buddha statues [31]). Experiments have additionally shown that our method can help mitigate some level of imbalance in the data. To further address this issue of imbalanced data, we further plan to study different strategies for pseudo-label assignment (such as incremental assignment) and the use of different loss functions in the GCN (such as the focal loss [21]).

These results open perspectives for tapping into the richness that unlabeled data can provide, and improve automatic art analysis. This should guide our future work towards zero-shot learning for automatic art analysis.

## Acknowledgments

This work is partly supported by JSPS KAKENHI Grant Numbers JP18H03571, JP18H03264, and JP20K19822.